%% file: main.tex
\documentclass[10pt,twocolumn,letterpaper]{article}

\usepackage[pagenumbers]{cvpr} % To force page numbers, e.g. for an arXiv version

\input{preamble}
\definecolor{cvprblue}{rgb}{0.21,0.49,0.74}
\usepackage[pagebackref,breaklinks,colorlinks,allcolors=cvprblue]{hyperref}
\usepackage{amsmath}
\usepackage{graphicx}
\usepackage{algorithm}
\usepackage{algorithmic}
\usepackage{multirow} 
\usepackage{amssymb}
\usepackage{bbding}
\usepackage{cuted}
\usepackage{caption}
\usepackage[utf8]{inputenc} % allow utf-8 input
\usepackage[T1]{fontenc}    % use 8-bit T1 fonts
\usepackage{hyperref}       % hyperlinks
\usepackage{url}            % simple URL typesetting
\usepackage{booktabs}       % professional-quality tables
\usepackage{amsfonts}       % blackboard math symbols
\usepackage{nicefrac}       % compact symbols for 1/2, etc.
\usepackage{microtype}      % microtypography
\usepackage{xcolor}         % colors
\usepackage{tabularx}
\usepackage{makecell}
\newcolumntype{C}{>{\centering\arraybackslash}X}
\usepackage{enumitem}

\def\confName{CVPR}
\def\confYear{2026}

\title{InvCoSS: Inversion-driven Continual Self-supervised Learning in Medical Multi-modal Image Pre-training}

\author{
Zihao Luo$^{1,2*}$, Shaohao Rui$^{2,3*}$, Zhenyu Tang$^{3,4}$, Guotai Wang$^{1,4,5}$\Envelope, Xiaosong Wang$^{2,4}$\Envelope\\
$^1$\normalsize University of Electronic Science and Technology of China, Chengdu, China \\
$^2$\normalsize Shanghai Innovation Institute, Shanghai, China 
$^3$\normalsize Shanghai Jiao Tong University, Shanghai, China \\
$^4$\normalsize Shanghai Artificial Intelligence Laboratory, Shanghai, China \\
$^5$\normalsize Brain-Computer Interface \& Brain-Inspired Intelligence Key Laboratory of Sichuan Province, Chengdu, China
\\
{\tt\small guotai.wang@uestc.edu.cn, wangxiaosong@pjlab.org.cn}\\
\href{https://zihaoluoh.github.io/InvCoSS}
{\texttt{https://zihaoluoh.github.io/InvCoSS}}
}

\begin{document}
\maketitle
\begin{strip}
    \centering
    \vspace*{-3.5em}
    \includegraphics[width=\textwidth]{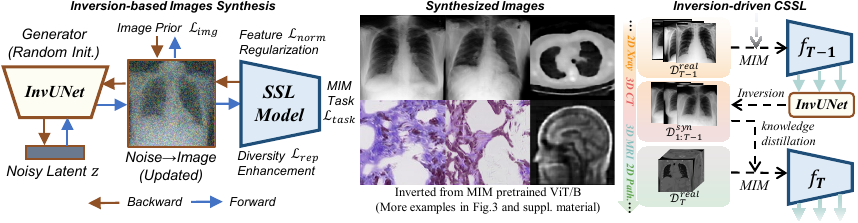} 
    \captionof{figure}{We propose InvCoSS, a novel inversion-driven continual self-supervised learning (CSSL) framework that utilizes synthetic images inverted from just self-supervised models as continual knowledge retention.
    % multi-modal medical images directly from pre-trained model parameters and normalization statistics to replace the real data buffer in conventional replay-based methods. 
    Besides, we design InvUNet (Sec.\ref{sec:invunet}) with multi-scale fusion to preserve high-frequency details, and introduce repulsive representation learning (Sec.\ref{sec:rrl}) to explicitly enforce diversity and prevent mode collapse, alongside feature regularization, task objectives, and image prior(Sec.\ref{sec:ooo}). Using these synthetic images for CSSL, we effectively mitigate catastrophic forgetting while eliminating data privacy constraints across institutes.}
    \label{fig:teaser}
\end{strip}
\input{sec/0_abstract}    
\input{sec/1_intro}
\input{sec/2_related_works}
\input{sec/3_methods}

\input{sec/4_experiments}

\input{sec/5_conclusion}

{
    \small
    \bibliographystyle{ieeenat_fullname}
    \bibliography{main}
}

\input{sec/appendix}

\end{document}

%% file: sec/0_abstract.tex
\begin{abstract}
Continual self-supervised learning (CSSL) in medical imaging trains a foundation model sequentially, alleviating the need for collecting multi-modal images for joint training and offering promising improvements in downstream performance while preserving data privacy. However, most existing methods still rely on replaying data from previous stages to prevent catastrophic forgetting, which compromises privacy and limits their applicability in real-world scenarios where data transfer across sites is often restricted. In this work, we propose \textbf{InvCoSS}, an \textbf{inv}ersion-driven \textbf{co}ntinual \textbf{s}elf-\textbf{s}upervised learning framework for medical multi-modal image pre-training. Specifically, after training on a previous task, InvCoSS inverts the pre-trained self-supervised model to generate synthetic images that approximate the original training distribution. These synthetic images are then combined with data from the new task for joint optimization, which effectively mitigates catastrophic forgetting while strictly adhering to the constraint of no access to previous real data. Furthermore, to improve the fidelity of synthetic images, we introduce a novel InvUNet with a multi-scale fusion architecture to restore both high- and low-frequency components of the inverted images. To enhance diversity and prevent mode collapse, we design a repulsive representation-learning mechanism that encourages a diverse feature space for synthetic images without class guidance. Extensive experiments across nine downstream tasks validate the effectiveness of InvCoSS, achieving performance comparable to or even superior to prior data-replay methods while significantly reducing storage requirements and eliminating data privacy constraints.
% \footnote{Code is available in the supplemental materials.}
\renewcommand{\thefootnote}{} 
\footnotetext{
\begin{minipage}[t]{\textwidth} 
$*$~Equal contribution\\
\Envelope~Corresponding authors
\end{minipage}}
\end{abstract}

%% file: sec/1_intro.tex
\section{Introduction}
\label{sec:intro}

Self-supervised learning (SSL) has substantially advanced unsupervised visual representation learning and has become the mainstream pre-training framework in medical image analysis, achieving generalization to diverse downstream tasks by learning from large-scale unlabeled data~\cite{ye2024continual,wu2024voco,he2023geometric,rui2025multi}. However, current medical self-supervised learning faces two primary challenges: (1) Limited to single modality pre-training: medical self-supervised learning methods are predominantly confined to single-modal pre-training, leading to a proliferation of single-modal SSL models, such as those for CT~\cite{haghighi2021transferable,yan2023localized}, X-ray~\cite{cai2023dual,zhou2021preservational,xiao2023delving}, and MR~\cite{tang2022self,valanarasu2023disruptive}. However, these single-modal self-supervised models frequently exhibit limited generalization capabilities across diverse modalities~\cite{ye2024continual}. This issue is often linked to the challenges of large-scale real-world data collection, which is hampered by ethical and privacy considerations~\cite{fini2020online,lopez2017gradient,qu2021recent}. Consequently, acquiring multi-modal image datasets for joint training becomes resource-intensive, thereby impeding the development of a unified multi-modal visual foundation model. (2) Limitations of static training paradigms: A further limitation inherent in existing SSL methods is their static training nature. This paradigm, prevalent in both single-modal and multi-modal SSL, typically relies on offline training on fixed datasets~\cite{fini2022self,wang2024comprehensive}. Consequently, integrating new data necessitates retraining on the entire updated dataset, which is highly impractical and prohibitively expensive given the substantial computational costs of SSL, thereby hindering model scalability. Moreover, one-time joint training on collected multi-modal data can lead to modality conflict, as observed in MedCoSS~\cite{ye2024continual}, where interference between modalities during training can detrimentally compromise the model's generalization capability.

A promising approach to address the aforementioned challenges is to train a unified multi-modal self-supervised model sequentially by enabling the model itself to function as a continual learner~\cite{fini2022self,ye2024continual}. In this paradigm, data is acquired progressively, and knowledge is incrementally integrated into the model, thereby alleviating the need for repeated joint training and eliminating the burden of data transfer due to privacy concerns. However, a notorious issue in continual learning is catastrophic forgetting, where knowledge acquired in previous stages is easily forgotten when adapting to new data distributions. One long-standing and effective mitigation strategy is data replay~\cite{ye2024continual}, which involves rehearsing subsets of previous training data during the current training stage to prevent forgetting. Although intuitive and practical, this method becomes infeasible when previous data is unavailable due to data privacy and ethical constraints. Hence, a critical question emerges: how can we preserve knowledge, as encapsulated in the model parameters, to the greatest extent without any replay data? Moreover, is it possible to preserve knowledge in a continual manner solely from the trained model itself?

Motivated by this, we draw inspiration from model inversion~\cite{yin2020dreaming,smith2021always,fang2021contrastive} in supervised learning, which aims to reconstruct the training data distribution solely from a well-trained model's checkpoints. We explore the feasibility of adapting such an approach for continual self-supervised learning. To this end, we propose InvCoSS, a novel inversion-driven continual self-supervised learning framework designed for medical multi-modal image pre-training. InvCoSS first generates synthetic images from the previous stage by leveraging the model's parameters and the statistical information stored in normalization layers. Subsequently, it utilizes these synthesized images to perform both standard feature distillation and task-oriented self-supervision, thereby effectively mitigating catastrophic forgetting. However, directly transferring model inversion techniques from supervised to self-supervised learning is highly non-trivial and presents several critical challenges. First, high-frequency details are often lost in inverted images, as illustrated in Fig.~\ref{fig:ablation_vis}. To address this issue, we introduce a novel InvUNet module, inspired by multi-scale fusion networks like U-Net~\cite{ronneberger2015u}, to enhance both high- and low-frequency information in the synthetic images. Second, the absence of explicit class guidance in self-supervised learning leads to aliasing and redundancy in the inverted images. To resolve this, we design a repulsive representation regularization module that explicitly enforces feature diversity in the representation space. We conduct extensive experiments across four medical imaging modalities and validate our approach on nine downstream tasks. The results show that InvCoSS achieves performance comparable to, and at times superior than, conventional data-replay methods, while completely avoiding raw data storage and privacy risks. By only retaining model parameters and normalization statistics, our approach significantly reduces storage costs and enhances practical applicability under real-world data transfer constraints. Our major contributions are four-fold:
\begin{itemize}
    \item We propose InvCoSS, a novel inversion-driven continual self-supervised learning framework for medical multi-modal image pre-training that mitigates catastrophic forgetting without requiring storage of any raw data.
    \item We design a novel InvUNet module that leverages a multi-scale fusion architecture that addresses high-frequency detail degradation in model inversion, enabling high-fidelity medical image synthesis across 2D and 3D modalities.
    \item We introduce repulsive representation learning that explicitly enforces feature diversity in the representation space, effectively preventing mode collapse in self-supervised inversion without class guidance.
    \item Extensive experiments across four medical imaging modalities and nine downstream tasks demonstrate that InvCoSS achieves performance on par with or superior to data-replay methods while reducing storage overhead by up to 590$\times$ and ensuring privacy preservation.
\end{itemize}

     % We propose InvCoSS, the first inversion-driven continual self-supervised learning framework for medical multi-modal pre-training that effectively mitigates catastrophic forgetting without requiring storage of any real data.
    
     %  We design a novel InvUNet module that leverages a multi-scale fusion architecture to address the challenge of high-frequency detail degradation in inverted images, significantly improving reconstruction quality.
     
     %  We introduce a contrastive feature distillation mechanism that explicitly enforces diversity in the synthesized feature space, effectively resolving the feature aliasing and redundancy issues in self-supervised inversion.
     
     %  We conduct extensive experiments demonstrating that our approach achieves performance comparable to or even superior than real data replay, while reducing storage requirements by over 99.95\% and completely eliminating privacy concerns.

%% file: sec/2_related_works.tex
\section{Related Work}
\label{sec:related_works}
\noindent{\textbf{Continual Self-supervised Learning (CSSL).}} CSSL refers to the progressive training of self-supervised models in an online manner, effectively addressing privacy, security, and proprietary concerns in real-world applications, particularly within the medical domain. Furthermore, CSSL has demonstrated enhanced effectiveness in handling multi-domain shifts \cite{ye2024continual,fini2022self}. Despite these advantages, significant research efforts focus on mitigating catastrophic forgetting in continual learning. Studies such as \cite{hu2021well,ye2024continual} have shown that straightforward continual learning strategies, including data replay and parameter regularization, can effectively alleviate knowledge forgetting. \cite{yao2025continual} investigated maintaining generalization performance in cross-modal continual learning, extending to vision-language modalities. \cite{tasai2025continual} explored multi-domain continual learning within the same modality by leveraging domain-invariant characteristics to constrain knowledge. \cite{tasai2025privacy} proposed retaining features from previous tasks instead of raw data to reduce privacy leakage. Collectively, these medical imaging continual self-supervised learning approaches typically employ data replay methods, storing past raw images or features and reusing them during subsequent pre-training stages to mitigate catastrophic forgetting. Although data replay has proven effective in both CL and CSSL scenarios, its implementation in medical settings remains impractical due to patient privacy concerns and ethical review requirements. To address this fundamental limitation, we investigate how to achieve optimal knowledge retention from previous pre-training stages under the most constrained conditions, utilizing solely the information obtained during pre-training and forgoing any reliance on auxiliary generative models (e.g., GANs~\cite{goodfellow2014generative}, Diffusion~\cite{ho2020denoising}) or dataset distillation methods~\cite{wang2018dataset}.

\noindent{\textbf{Model inversion in Knowledge Distillation.}} 
Model inversion originated from the technique introduced in~\cite{mordvintsev2015inceptionism}, which demonstrates how to synthesize images by optimizing input noise through gradient descent in a classification network, given output classes and image priors such as correlated neighboring pixels. It has been extensively applied in data-free knowledge distillation (DF-KD), where knowledge is transferred from a teacher model to a student model without accessing the original training data, following an inversion-and-distillation paradigm~\cite{liu2024small}. This process first transforms a noise distribution into a training data distribution guided by the well-trained teacher model, and then performs knowledge distillation on these synthesized data. \cite{yin2020dreaming} pioneered the use of Batch Normalization layer parameters for feature distribution regularization, significantly enhancing the fidelity of inverted images without requiring additional specialized training, as these parameters are naturally obtained after model training. Subsequent research has focused on incorporating more inversion priors and improving inversion speed. For example, a line of generator-based methods~\cite{chen2019data,yoo2019knowledge,li2023dynamic} accelerates model inversion by reusing generators to transform noisy latents to high-dimensional images. Another line of work aims to enhance the efficiency of inverted images for distillation through meta-learning~\cite{fang2022up,patel2023learning} or adversarial learning~\cite{do2022momentum,yu2023data}. Overall, these model inversion methods for knowledge distillation are limited in the supervised learning framework (e.g., image classification). Distinct from prior arts, our work is the first to explore model inversion for self-supervised learning which presents unique challenges due to the absence of explicit supervisory signals. To address the high-frequency degradation inherent in SSL inversion, we design a novel InvUNet module with a multi-scale fusion architecture. Furthermore, to compensate for the lack of class guidance, we introduce a repulsive representation learning mechanism that explicitly enforces feature diversity in the feature space.
% Importantly, we maintain the data-free nature by not explicitly training a generative network during previous stages of pre-training, relying instead on the core idea of \cite{yin2020dreaming}: leveraging model parameters and normalization layer statistics to synthesis images from that stage for subsequent continual learning.

%% file: sec/3_methods.tex
\section{Methods}
\label{sec:methods}
\label{sec:synthesis}
\begin{figure*}
    \centering
    \includegraphics[width=1\textwidth]{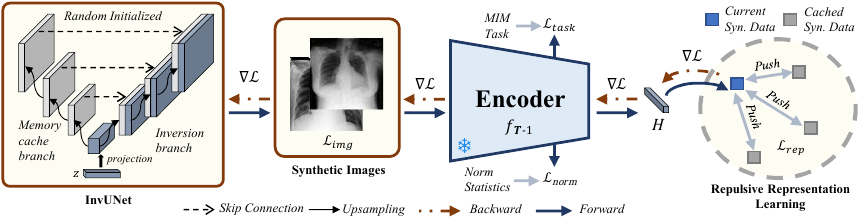}
    \caption{Overview of inversion-based image synthesis. InvUNet synthesizes images from a bottleneck injected latent $z$, with a memory-cache branch providing multi-scale priors via skip connections. A frozen model $f_{T-1}$ supervises with MIM task prior $\mathcal{L}_{\mathrm{task}}$, norm statistics matching $\mathcal{L}_{\mathrm{norm}}$, and total variation image prior $\mathcal{L}_{\mathrm{img}}$, while a persistent feature pool imposes a repulsive loss $\mathcal{L}_{\mathrm{rep}}$ to promote diversity. The process is data-free and produces synthetic datasets $D_t^{syn}$ for knowledge retention at stage T.
    }
    \label{fig:inversionpipline}
\end{figure*}
% We introduce \textbf{I}nversion-driven \textbf{Co}ntinual \textbf{S}elf-\textbf{S}upervised Learning (InvCoSS), a data-free continual knowledge distillation framework that addresses the critical privacy and storage challenges which prohibit raw data retention in applications like multi-center medical studies. 

% It circumvents this critical bottleneck by replacing raw data replay with an inverted generative approach conditioned on data statistics.

\subsection{Preliminaries}
\noindent{\textbf{Model inversion.}} Given a well-trained model $f$, model inversion aims to synthesize images that approximate the training data distribution. Formally, given a learnable input noisy image $\hat{x}$, the synthesized image $x^{syn}$ is obtained by optimizing the following objective:
\begin{equation}
\label{eq:inversion_loss}
    x^{syn} = \arg\min_{\hat{x}} \left[ \mathcal{L}_{\mathrm{task}}(\hat{x};f) + \mathcal{R}(\hat{x};f) \right],
\end{equation}
where $\mathcal{L}_{\mathrm{task}}(\cdot)$ denotes the task-oriented loss (e.g., cross-entropy loss in classification), and $\mathcal{R}(\cdot)$ represents the image regularization term which usually incorperating image priors~\cite{mordvintsev2015inceptionism} and feature distribution regularization~\cite{yin2020dreaming}:
\begin{equation}
    \mathcal{R}(\hat{x}) = \mathcal{L}_{\mathrm{img}}(\hat{x}) + \mathcal{L}_{\mathrm{norm}}(\hat{x};f). 
    % + \mathcal{L}_{\mathrm{rep}}(z).
\end{equation}
The feature regularization term $\mathcal{L}_\mathrm{norm}$ specifically enforces feature map distribution match for $x^{syn}$ and $x^{real}$ by assuming its Gaussianity across batches~\cite{yin2020dreaming}:
\begin{equation}
\mathcal{L}_{\mathrm{norm}}(\hat{x}) = 
\sum_{l=1}^{L} \left( 
\left\| \mu_l(\hat{x}) - \mathbb{E}[\mu_l] \right\|_2 
+ 
\left\| \sigma_l^2(\hat{x}) - \mathbb{E}[\sigma_l^2] \right\|_2 
\right),
\label{eq:bn_loss}
\end{equation}
where $\mu_l(\cdot)$ and $\sigma_l(\cdot)$ compute the batch-wise mean and variance of feature activations at the $l$-th layer, respectively. The expectation terms $\mathbb{E}[\mu_l]$ and $\mathbb{E}[\sigma_l^2]$ are approximated using the running averages maintained during training.

\noindent{\textbf{Continual Self-supervised Learning.}} We consider a data replay based standard CSSL scenario following MedCoSS~\cite{ye2024continual} composed of a sequence of $T$ tasks. Each task $t$ provides an unlabeled dataset $\mathcal{D}_t$ drawn from distinct imaging modalities. The model $f_t$ for current task $t$ is trained by optimizing the following objective:
\begin{equation}
\label{eq:ssl_loss}
f_t = \arg\min_{f} \left[ \mathcal{L}_{\mathrm{task}}(f, \mathcal{D}_t) + \mathcal{L}_{\mathrm{reg}}(f, f_{t-1}; \mathcal{D}_{\mathrm{buff}}) \right],
\end{equation}
where we use the self-supervised learning objective as $\mathcal{L}_{\mathrm{task}}$ (see Eq.~\ref{eq:loss_task}). $\mathcal{L}_{\mathrm{reg}}$ is a regularization term that uses the previous model $f_{t-1}$ and a replay buffer $\mathcal{D}_{\mathrm{buff}}$ of stored samples from earlier tasks to alleviate catastrophic forgetting. In data-free settings, $\mathcal{D}_{\mathrm{buff}}$ is either empty or substituted with synthetic data generated solely from the knowledge within $f_{t-1}$ (our method), which introduces further challenges for preserving past knowledge without access to raw data.
% Our approach fundamentally departs from replay-based methods that store raw samples in a buffer for replay. Instead of maintaining such a buffer, our framework discards the raw data after training on task $t$ and retains only a compact set of Global Latent Statistics (GLS), denoted as $\mathcal{S}^{\text{GLS}}_t$. These statistics serve as a compressed, privacy-preserving representation of the task's data distribution.
\subsection{InvCoSS}
\label{sec:overview}
At self-supervised learning stage $T$, our objective is to effectively learn from the newly arrived dataset $\mathcal{D}_T$ while preserving knowledge acquired from previous tasks $\{1, \dots, T-1\}$. To achieve this under data-free constraints, we employ model inversion techniques to synthesize images for knowledge retention. Specifically, we adopt generator-based model inversion methods~\cite{fang2021contrastive,liu2024small} that map low-dimensional noisy latents to high-dimensional image data, thereby minimizing computational overhead. For each previous task $t \in \{1, \dots, T-1\}$, we generate synthetic image data by optimizing the inversion objective in Eq.~\ref{eq:inversion_loss}, yielding synthetic datasets $\mathcal{D}^{syn}_t = \{x_{t_i}^{syn}\}_{i=1}^{N_t}$ where each $x_{t_{i}}^{syn}$ is reconstructed exclusively from the pre-trained model of task $t$. Detailed CSSL paradigms are in on Appendix \ref{app:invcoss}.

The synthetic datasets are aggregated into a comprehensive sample pool $\mathcal{B}_T = \bigcup_{t=1}^{T-1} \mathcal{D}_t^{syn}$. The model $f_T$ is then trained using both the current task data $\mathcal{D}_T$ and the synthetic sample pool $\mathcal{B}_T$ through the joint self-supervised objective defined in Eq.~\ref{eq:ssl_loss}. While this training strategy aligns with the MedCoSS~\cite{ye2024continual}, the key distinction lies in our utilization of synthetic samples $\mathcal{B}_T$ as a substitute for raw data in the buffer. This approach enables effective learning from current task data while preserving knowledge from previous stages through synthetic samples, thereby mitigating catastrophic forgetting within a privacy-preserving and storage-efficient paradigm. 

\subsection{InvUNet for High-fidelity Image Synthesis} 
\label{sec:invunet}
% The core of our data-free replay in InvCoSS is a generator-based synthesis mechanism that produces high-fidelity medical images representative of a previously trained model, $f_{T-1}$. This approach learns a shared, generative mapping from a latent space, circumventing the impracticality of optimizing individual images at the pixel level. To this end, we propose InvUNet, a generator $\mathcal{G}$ that maps a trainable latent vector $z$ to a synthetic image ${x}^{syn}_t = \mathcal{G}(z)$. The objective is to optimize the parameters of $\mathcal{G}$ such that its output distribution effectively mimics the one captured by $f_{t-1}$.

% \noindent\textbf{Generator Architecture.} 
% Our InvUNet generator is specifically designed to address a key challenge in generative modeling: the degradation of high-frequency details that leads to blurry or unrealistic synthetic images. While a U-Net structure~\cite{ronneberger2015u} excels at multi-scale fusion, adapting it for generation by directly projecting a low-dimensional latent vector into a high-resolution feature space is a notoriously difficult task.
During our pilot experiments on model inversion for self-supervised learning models, we observe that direct bottom-to-top projection of low-dimensional noisy vectors into high-dimensional images causes severe degradation of high-frequency details, yielding blurry and distorted synthetic images as shown in the third row of Fig.~\ref {fig:ablation_vis}. To overcome this issue, we propose InvUNet, a novel dual-stream generator network inspired by the U-Net architecture~\cite{ronneberger2015u} designed for multi-scale feature fusion. As illustrated in the left part of Fig.~\ref{fig:inversionpipline}, InvUNet injects the noisy latent $z$ directly into the network bottleneck, creating an information bottleneck that forces $z$ to encapsulate essential semantic information. This design enables a specialized division of labor between two complementary branches: the lightweight Memory Cache Branch generates multi-scale structural priors, while the primary Inversion Branch concentrates exclusively on high-fidelity semantically-guided inversion. Skip connections between these branches establish vital gradient pathways that streamline optimization by propagating error signals back to the Memory Cache Branch, thereby ensuring effective recovery of fine-grained details. Complete architectural specifications are provided in the Appendix~\ref{app:invunet}.

\subsection{Repulsive Representation Learning} 
\label{sec:rrl}
To mitigate mode collapse~\cite{metz2017unrolled} and enhance sample diversity, we introduce a repulsive representation learning mechanism.
% While the preceding losses ($\mathcal{L}_{\mathrm{task}}$, $\mathcal{L}_{\mathrm{norm}}$, $\mathcal{L}_{\mathrm{img}}$) ensure synthetic fidelity and statistical alignment, they do not inherently guarantee diversity. 
Specifically, we explicitly maximizes diversity in representation space by maintaining a persistent synthetic data pool $\mathcal{P}$ with size matching the number of synthetic samples to be generated.  For a batch of synthetic images $\mathcal{X}^{syn}=\{x^{syn}_i\}_{i=1}^B$, we extract features $h_i=f_{enc}\left(x^{syn}_i\right)$ using the frozen encoder $f_{enc}$ of $f_{T-1}$, the loss minimizes their cosine similarity to all features in $\mathcal{P}$:
\begin{equation}
\label{eq:rep_loss_pool}
\mathcal{L}_{\mathrm{rep}}(\mathcal{X}^{syn}, \mathcal{P};f_{T-1}) = \frac{1}{B \cdot |\mathcal P|} \sum_{i=1}^{B} \sum_{j=1}^{|\mathcal P|} \left( \frac{h_i \cdot p_j}{\|h_i\|_2 \|p_j\|_2} \right)^2.
\end{equation}
% Where $\hat{P}$ is the current number of feature in $\mathcal{P}$.
This purely repulsive formulation, typically unstable in standard contrastive learning without positive pairs, is uniquely stabilized in our framework by the complementary objectives, which act as semantic anchors to constrain synthesis within the learned data distribution.
\subsection{Overall Inversion Objective} 
\label{sec:ooo}
Fig.~\ref{fig:inversionpipline} illustrates our inversion process, which jointly optimizes the parameters $\theta_G$ of a randomly initialized InvUNet $\mathcal{G}$ and a batch of noisy latent variable $z$. The overall objective $\mathcal{L}_{\text{Inv}}$ is formulated as:
\begin{equation}
\label{eq:generator_loss}
\mathcal{L}_{\text{Inv}}(\theta_G, z) = \begin{aligned}[t]
    & \mathcal{L}_{\mathrm{task}}(\mathcal{G}(z;\theta_G); f_{T-1}) \\
    & + \alpha_{\mathrm{norm}}\mathcal{L}_{\mathrm{norm}}(\mathcal{G}(z;\theta_G)) \\
    & + \alpha_{\mathrm{img}} \mathcal{L}_{\mathrm{img}}(\mathcal{G}(z;\theta_G)) \\
    & + \alpha_{\mathrm{rep}} \mathcal{L}_{\mathrm{rep}}(\mathcal{G}(z;\theta_G); f_{T-1}),
\end{aligned}
\end{equation}
where $\alpha_{\mathrm{norm}}, \alpha_{\mathrm{img}}, \alpha_{\mathrm{rep}}$ are hyperparameters that balance each term. Each component is detailed below.

\noindent\textbf{Task-oriented Supervision.} Following MedCoSS~\cite{ye2024continual} training protocol, we employ masked image modeling~\cite{he2022masked} as the SSL objective. For each synthetic image $\mathcal{G}(z;\theta_G)$ (simply denoted ${x}^{syn}_{t_{i}}$), the task-oriented loss computes the reconstruction error within the masked regions $M$:
\begin{equation}
    \label{eq:loss_task}
    \mathcal{L}_{\mathrm{task}}({x}^{syn}_{t_{i}};f_{T-1}) = \left\| \left(f_{T-1}({x}^{syn}_{t_{i}} \odot \bar{M}) - {x}^{syn}_{t_{i}} \right) \odot M \right\|_2^2,
\end{equation}
where $\bar{M} = 1-M$ and $\odot$ is element-wise multiplication. 

\noindent\textbf{Feature Distribution Regularization.} We employ the normalization statistics loss $\mathcal{L}_{\mathrm{norm}}$ from Eq.~\ref{eq:bn_loss} to align the feature distribution of synthetic images with the original data by matching their batch-wise statistics in $f_{T-1}$'s normalization layers. Details of the normalization statistics calculation are provided in the Appendix~\ref{app:mi}.

\noindent\textbf{Image Prior Loss.} We employ a total variation loss as an image prior to promote spatial smoothness and suppress high-frequency artifacts. For a 2D synthetic image ${x}^{syn}_t \in \mathbb{R}^{H \times W}$, it penalizes adjacent pixel differences:
\begin{equation}
    \mathcal{L}_{\mathrm{img}}({x}^{syn}_t) = \sum_{i,j} \left( |x_{i+1,j} - x_{i,j}| + |x_{i,j+1} - x_{i,j}| \right).
\end{equation}
where $i,j$ denote spatial indices. For 3D volumes ${x}^{syn}_t \in \mathbb{R}^{D \times H \times W}$, this extends to include depth-wise differences.

%% file: sec/4_experiments.tex
\section{Experiments and Results}
\begin{table*}[!t]
   \centering
   \caption{Performance comparison on nine downstream medical tasks. The AVG$\uparrow$ and AVG$\downarrow$ are averaged across all metrics, the higher score of which indicates better and worse performance, respectively. For each metric, the \textbf{best} and \underline{second-best} results are highlighted.}
   \label{tab:main_results}
   \resizebox{\textwidth}{!}{%
   \small
   \setlength{\tabcolsep}{2pt}
   \begin{tabular}{@{}l ccc| ccc| cc| ccc| cc| cc| cc| ccc| cc| cc@{}}
       \toprule
       \multirow{2}{*}{\textbf{Method}} & \multicolumn{3}{c}{PudMed20k} & \multicolumn{3}{c}{ChestXR} & \multicolumn{2}{c}{QaTa} & \multicolumn{3}{c}{RICORD} & \multicolumn{2}{c}{LiTS} & \multicolumn{2}{c}{VS} & \multicolumn{2}{c}{LA} & \multicolumn{3}{c}{NCH} & \multicolumn{2}{c}{GlaS} & \multicolumn{2}{c}{Average} \\
       \cmidrule(lr){2-4} \cmidrule(lr){5-7} \cmidrule(lr){8-9} \cmidrule(lr){10-12} \cmidrule(lr){13-14} \cmidrule(lr){15-16} \cmidrule(lr){17-18} \cmidrule(lr){19-21} \cmidrule(lr){22-23} \cmidrule(lr){24-25}
       & ACC & AUC & F1 & ACC & AUC & F1 & DSC & HD & ACC & AUC & F1 & DSC & HD & DSC & HD & DSC& HD & ACC & AUC & F1 & DSC & HD & AVG$\uparrow$ & AVG$\downarrow$ \\
       \midrule
       \multicolumn{25}{c}{\textit{Static Pre-training Baselines}} \\
       \midrule
       TFS & 82.41 & 94.82 & 76.45 & 82.36 & 94.32 & 81.55 & 73.94 & 38.38 & 73.02 & 79.69 & 81.25 & 60.77 & 67.40 & 84.27 & 58.15 & 85.94 & 16.62 & 84.93 & 97.37 & 79.15 & 86.20 & 51.88 & 82.26 & 46.49 \\
       Report & 83.79 & 95.33 & 77.97 & 79.73 & 92.67 & 79.04 & 72.31 & 40.29 & 74.21 & 81.42 & 82.36 & 59.71 & 63.41 & 43.17 & 151.83 & 83.92 & 17.45 & 86.31 & 97.65 & 81.07 & 83.75 & 56.12 & 79.67 & 65.82 \\
       X-ray & 82.40 & 94.80 & 76.55 & \textbf{95.55} & \textbf{99.21} & \textbf{95.08} & \textbf{80.27} & \textbf{27.47} & 76.98 & 84.77 & 83.28 & 62.96 & 55.79 & 87.07 & 26.86 & 87.90 & 10.85 & 93.26 & 99.04 & 90.57 & 86.45 & 47.82 & 86.83 & 33.76 \\
       CT & 82.59 & 94.91 & 76.84 & 85.37 & 95.63 & 84.69 & 74.40 & 36.10 & 80.56 & 85.74 & 86.09 & 71.98 & \underline{33.79} & 89.90 & 10.91 & 87.21 & 16.01 & 88.25 & 98.31 & 83.79 & 84.52 & 50.81 & 85.34 & 29.52 \\
       MRI & 82.82 & 94.93 & 76.96 & 84.14 & 95.14 & 83.47 & 74.08 & 36.22 & 79.37 & 87.60 & 85.11 & 68.54 & 43.79 & 89.83 & 12.07 & 89.64 & \underline{10.24} & 89.28 & 99.43 & 83.72 & 83.52 & 53.93 & 85.15 & 31.25 \\
       Path. & 82.15 & 94.71 & 76.22 & 92.13 & 98.35 & 91.45 & 78.56 & 30.66 & 73.41 & 82.30 & 80.68 & 61.78 & 57.97 & 85.26 & 46.74 & 88.93 & 11.23 & 95.62 & \underline{99.54} & \textbf{94.01} & \underline{89.51} & 46.79 & 86.15 & 38.68 \\
       Joint SSL* & \textbf{84.10} & \textbf{95.52} & \textbf{78.30} & 92.24 & {98.33} & 91.59 & 78.61 & 30.29 & 77.78 & 85.89 & 83.19 & 70.59 & 38.24 & 89.73 & 15.33 & 89.41 & 13.84 & 94.63 & 99.27 & 92.93 & 87.86 & 51.20 & 87.65 & 29.78 \\
       Joint SSL† & \underline{83.94} & 95.33 & 78.13 & 91.54 & 98.02 & 90.92 & 78.29 & 30.27 & 78.97 & 86.95 & 84.00 & 69.60 & 44.17 & 89.47 & 18.31 & 88.44 & 13.84 & 94.44 & 99.30 & 92.69 & 87.83 & 52.08 & 87.52 & 31.73 \\
       \midrule
       \multicolumn{25}{c}{\textit{Continual Self-supervised Learning}} \\
       \midrule
       \multicolumn{25}{l}{\textit{Data-replay Methods}} \\
       ER\cite{riemer2018learning} & 83.75 & \underline{95.43} & \underline{78.15} & 85.32 & 95.12 & 84.43 & 75.03 & 36.59 & 75.79 & 83.51 & 82.03 & 69.53 & 41.86 & 88.61 & 25.47 & 87.37 & 14.42 & 92.21 & 99.14 & 89.54 & 87.61 & 53.10 & 85.45 & 34.29 \\
       MedCoSS\cite{ye2024continual} & 83.59 & 95.38 & 77.87 & \underline{94.31} & \underline{98.83} & \underline{93.77} & \underline{78.98} & 29.43 & \underline{83.33} & \underline{88.74} & \underline{87.87} & \underline{72.01} & 36.50 & \textbf{90.12} & \textbf{7.80} & \textbf{90.46} & \textbf{9.55} & \textbf{95.76} & 99.51 & \textbf{94.01} & 89.13 & \underline{46.69} & \underline{89.03} & \underline{25.99} \\
       \midrule
       \multicolumn{25}{l}{\textit{Data-free Methods}} \\
       EWC\cite{kirkpatrick2017overcoming} & 83.39 & 95.17 & 77.66 & 90.48 & 97.77 & 89.72 & 77.44 & 31.67 & 75.79 & 82.18 & 82.59 & 64.59 & 53.66 & 88.15 & 23.87 & 88.00 & 12.55 & 94.50 & 99.37 & 92.61 & 88.16 & 50.81 & 86.27 & 34.51 \\
       PackNet\cite{mallya2018packnet} & 82.88 & 95.01 & 77.06 & 86.14 & 95.98 & 85.28 & 75.69 & 35.98 & 71.43 & 83.75 & 79.63 & 61.89 & 62.12 & 83.92 & 54.07 & 88.87 & 12.55 & 92.42 & 99.13 & 90.00 & 86.96 & 52.81 & 84.42 & 43.51 \\
       CaSSLe\cite{fini2022self} & 82.93 & 95.12 & 77.04 & 91.66 & 98.07 & 90.99 & 78.19 & 31.23 & 76.59 & 82.85 & 84.32 & 67.04 & 47.17 & 87.94 & 21.47 & 89.66 & 10.67 & 95.01 & 99.44 & 93.06 & 89.12 & 47.93 & 86.88 & 31.69 \\
       \textbf{InvCoSS} & 83.63 & 95.24 & 77.81 & 94.09 & 98.52 & 93.53 & 78.70 & \underline{29.07} & \textbf{84.52} & \textbf{89.23} & \textbf{89.43} & \textbf{72.14} & \textbf{30.36} & \underline{89.93} & \underline{10.25} & \underline{90.21} & {10.76} & \underline{95.75} & \textbf{99.61} & \underline{93.90} & \textbf{89.56} & \textbf{45.24} & \textbf{89.17} & \textbf{25.14} \\
      
       \bottomrule
       \multicolumn{25}{l}{\textsuperscript{*}With a shared decoder. \textsuperscript{†}With separate decoders.} \\
   \end{tabular}
   }
\end{table*}
\begin{figure*}[htbp]
    \centering
    \includegraphics[width=\textwidth]{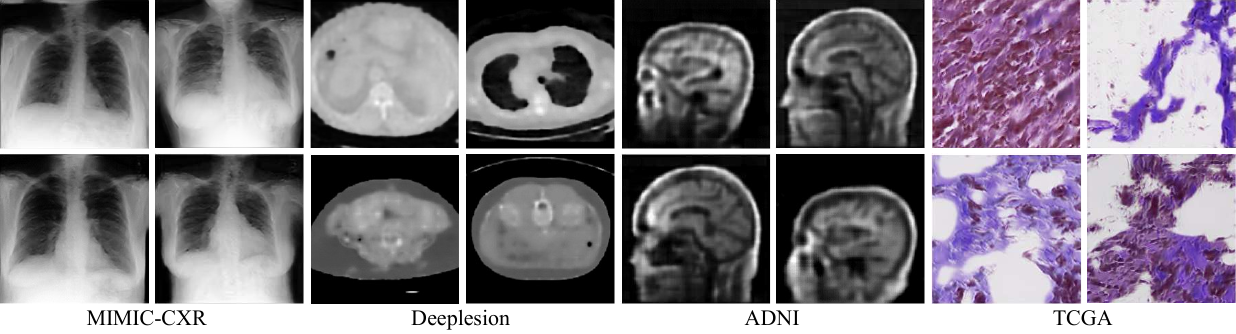}
    \caption{Visualizations of synthetic data generated by our inversion-driven framework, showcasing its versatility across diverse medical tasks. The figure displays samples from four imaging modalities (two columns each): X-ray, CT, MR, and Pathological imaging.}
    \label{fig:vis}
\end{figure*}
\subsection{Experimental Setups}
\noindent{\textbf{Datasets.}} We follow the experimental setup and dataset selection from MedCoSS~\cite{ye2024continual}, utilizing a large-scale multi-modal pre-training corpus and evaluating on a diverse set of nine downstream benchmarks. The pre-training corpus is assembled from five unlabeled public datasets across distinct medical modalities: 1D clinical reports and 2D chest radiographs from MIMIC-CXR~\cite{johnson2019mimic}, 3D CT scans from DeepLesion~\cite{yan2018deeplesion}, 3D MRI scans from the ADNI database~\cite{jack2008alzheimer}, and histopathology images from 7 The Cancer Genome Atlas (TCGA) projects: TCGA-THYM, TCGA-THCA, TCGA-BRCA, TCGA-UCEC, TCGA-UVM, TCGA-OV, TCGA-MESO. Considering our model inversion method is designed for image modality, we keep the same real text buffer for fair comparison with MedCoSS~\cite{ye2024continual}. We evaluate our model on nine downstream benchmarks across five modalities: report classification on PubMed20k~\cite{dernoncourt2017pubmed}; X-ray analysis with COVID-19 classification on ChestXR~\cite{Akhloufi2021} and lesion segmentation on QaTa-COV19-v2~\cite{degerli2022osegnet}; CT analysis via COVID-19 classification on RICORD~\cite{tsai2021rsna} and liver and tumor segmentation on LiTS~\cite{bilic2023liver}; MRI segmentation on Vestibular Schwannoma (VS)~\cite{shapey2021segmentation} and Left Atrium (LA)~\cite{xiong2021global} datasets; and pathology analysis using NCH~\cite{kather2018100} and GlaS~\cite{sirinukunwattana2017gland} for tissue classification and gland segmentation, respectively. All data splits and pre-processing follow MedCoSS~\cite{ye2024continual}. Further dataset details are provided in Appendix~\ref{app:ds}.
% For downstream evaluation, we assess our model's performance on nine benchmarks spanning all five modalities. These tasks include report classification on PudMed20k~\cite{dernoncourt2017pubmed}; X-ray analysis via COVID-19 classification on ChestXR~\cite{Akhloufi2021} and lesion segmentation on QaTa-COV19-v2~\cite{degerli2022osegnet}; CT analysis through COVID-19 classification on RICORD~\cite{tsai2021rsna} and liver and liver tumor segmentation on LiTS~\cite{bilic2023liver}; MRI segmentation on the Vestibular Schwannoma (VS)~\cite{shapey2021segmentation} and Left Atrium (LA)~\cite{xiong2021global} datasets; and finally, pathology analysis using the NCH~\cite{kather2018100} and GlaS~\cite{sirinukunwattana2017gland} benchmarks for tissue classification and gland segmentation, respectively. All data splits and pre-processing protocols are identical to those used in MedCoSS~\cite{ye2024continual}. More details of pre-training and downstream datasets are in Appendix~\ref{app:pt} and Appendix~\ref{app:ds}.

\noindent\textbf{Implementation Details}
Following MedCoSS~\cite{ye2024continual}, we use the pre-training modality order: Report, X-ray, CT, MRI, and Pathological imaging. Input sizes are set to $112$ for 1D, $224\times224$ for 2D, and $16\times192\times192$ for 3D. For pre-training, we use the AdamW optimizer with a batch size of 512 and train for 300 epochs per modality. The learning rate warms up to $1.5\times10^{-4}$ over 40 epochs and follows a cosine decay.
We detail the computation of batch-wise statistics for ViT in Appendix~\ref{app:mi}. The InvUNet model is optimized using Adam, with learning rates set to $2\times10^{-4}$ for the generator and $0.05$ for the latent vectors~$z$. Loss weights are set to $\alpha_{\text{norm}}=1$, $\alpha_{\text{img}}=0.1$, and $\alpha_{\text{rep}}=0.1$. All downstream task configurations are identical to MedCoSS. Further details are provided in Appendix~\ref{app:id}.

\noindent\textbf{Comparison Methods.}
% Following the same experimental protocol as MedCoSS~\cite{ye2024continual}, we evaluate InvCoSS against their reported baselines across nine downstream medical tasks. The compared methods fall into two main categories: static pre-training baselines and continual self-supervised learning approaches. 
% The static pre-training baselines consist of training from scratch, five single-modal paradigms trained on clinical reports, X-rays, CT scans, MRI, and pathological images, respectively, and two joint SSL variants with either a shared decoder or separate decoders. 
% The CSSL methods are categorized into data-free approaches, including EWC~\cite{kirkpatrick2017overcoming}, PackNet~\cite{mallya2018packnet}, and CaSSLe~\cite{fini2022self}, and data-replay approaches, including ER~\cite{riemer2018learning} and MedCoSS~\cite{ye2024continual}. Data-replay methods maintain a buffer containing 5\% of previous task data, and InvCoSS correspondingly generates an equivalent number of synthetic images to ensure fair comparison. All methods utilized masked image/text modeling as the pretext task and ViT/B as the backbone. 
Following MedCoSS~\cite{ye2024continual}, we evaluate InvCoSS against their reported baselines across nine downstream medical tasks. Compared methods include static pre-training baselines and CSSL approaches. Static baselines comprise training from scratch, five single-modal paradigms (clinical reports, X-rays, CT, MRI, pathology), and two joint SSL variants with shared/separate decoders. CSSL methods include data-free approaches (EWC~\cite{kirkpatrick2017overcoming}, PackNet~\cite{mallya2018packnet}, CaSSLe~\cite{fini2022self}) and data-replay approaches (ER~\cite{riemer2018learning}, MedCoSS~\cite{ye2024continual}). Data-replay methods maintain a 5\% data buffer, while InvCoSS generates equivalent synthetic samples for fair comparison. All methods employ masked image/text modeling with ViT/B backbone.
\\
\noindent\textbf{Evaluation Metrics.}
For segmentation tasks, we adopt the Dice similarity coefficient (DSC) and the 95th percentile Hausdorff Distance (HD) as evaluation metrics. For classification tasks, we report Accuracy (ACC), Area Under the Curve (AUC), and F1 Score as performance measures. For robustness, we average results over three random seeds: 0, 10, and 100.
\subsection{Main Results}
\begin{table*}[!t]
   \centering
   \caption{Ablation study on the effectiveness of different components. The AVG$\uparrow$ and AVG$\downarrow$ indicate the average values across all metrics, where higher is better and lower is better, respectively. For each metric, the best results are highlighted with \textbf{bold}.}
   \label{tab:ablation}
   \resizebox{0.85\textwidth}{!}{%
   \small
   \setlength{\tabcolsep}{1.8pt}
   \begin{tabular}{@{}ccccc| ccc| cc| ccc| cc| cc| cc| cc@{}}
       \toprule
       \multicolumn{5}{c}{\textbf{Components}} & \multicolumn{3}{c}{ChestXR} & \multicolumn{2}{c}{QaTa} & \multicolumn{3}{c}{RICORD} & \multicolumn{2}{c}{LiTS} & \multicolumn{2}{c}{VS} & \multicolumn{2}{c}{LA} & \multicolumn{2}{c}{Average} \\
       \cmidrule(lr){1-5} \cmidrule(lr){6-8} \cmidrule(lr){9-10} \cmidrule(lr){11-13} \cmidrule(lr){14-15} \cmidrule(lr){16-17} \cmidrule(lr){18-19} \cmidrule(lr){20-21}
       $\mathcal{L}_{task}$ & $\mathcal{L}_{BN}$ & $\mathcal{L}_{img}$ & $\mathcal{L}_{rep}$ & $\mathcal{G}$ & ACC & AUC & F1 & DSC & HD & ACC & AUC & F1 & DSC & HD & DSC & HD & DSC & HD & AVG$\uparrow$ & AVG$\downarrow$ \\
       \midrule
       % & & & & & 79.73 & 92.67 & 79.04 & 72.31 & 40.29 & 74.21 & 81.42 & 82.36 & 59.71 & 63.41 & 43.17 & 151.83 & 83.92 & 17.45 & 71.03 & 68.25 \\
       \checkmark & \checkmark & & & \checkmark& 93.38 & \textbf{98.81} & 93.30 & \textbf{78.99} & \textbf{28.82} & 78.57 & 86.60 & 85.11 & 70.16 & 36.27 & 88.35 & 28.41 & 87.94 & 14.99 & 86.12 & 27.12 \\
       \checkmark & \checkmark & \checkmark & & \checkmark& 93.91 & 98.69 & 93.31 & 78.45 & 29.42 & 82.14 & 85.03 & 87.60 & 70.62 & 36.47 & 88.61 & 20.14 & 89.12 & 11.12 & 86.75 & 24.29 \\
       \checkmark & \checkmark & \checkmark & \checkmark & \checkmark & \textbf{94.09} & 98.52 & \textbf{93.53} & 78.70 & 29.07 & \textbf{84.52} & \textbf{89.23} & \textbf{89.43} & \textbf{72.14} & \textbf{30.36} & \textbf{89.93} & \textbf{10.25} & \textbf{90.21} & \textbf{10.76} & \textbf{88.03} & \textbf{20.11} \\
        \midrule
        \checkmark & \checkmark & \checkmark & \checkmark & & 92.24 & 97.63 & 91.59 & 76.21 & 39.72 & - & - & - & - & - & - & - & - & - & - & - \\
       \bottomrule
       \multicolumn{21}{l}{`-' indicates an OOM issue has occurred.}
   \end{tabular}
   }
\end{table*}
\label{sec:experiment}
\noindent\textbf{Quantitative Evaluation.} Table~\ref{tab:main_results} presents a comprehensive comparison across nine downstream tasks spanning five modalities. Joint SSL* achieves the best static baseline performance with 87.65 AVG$\uparrow$ and 29.78 AVG$\downarrow$ by accessing all modalities simultaneously, while single-modal methods excel on their respective domains but generalize poorly across modalities. InvCoSS substantially outperforms all data-free continual learning baselines, including EWC at 86.27 AVG$\uparrow$, PackNet at 84.42 AVG$\uparrow$, and CaSSLe at 86.88 AVG$\uparrow$, achieving 89.17 AVG$\uparrow$. More critically, InvCoSS matches or exceeds data-replay methods while maintaining complete privacy, surpassing ER by +3.72\% and achieving comparable performance to MedCoSS at 89.03 AVG$\uparrow$ and 25.99 AVG$\downarrow$ without storing any real data. Notably, InvCoSS demonstrates superior performance on CT tasks, achieving best results on RICORD at 84.52\% ACC and LiTS at 72.14\% DSC, outperforming MedCoSS by +1.19\% and +0.49\% on RICORD. This suggests that synthetic data generated via model inversion may better capture the original data distribution than clustered subsets used in replay methods.

\noindent\textbf{Qualitative Analysis.} Fig.~\ref{fig:vis} visualizes synthetic data generated across four medical imaging modalities. The inverted images exhibit anatomically plausible structures across X-ray, CT, MRI, and pathological imaging. While the synthetic samples lack fine-grained textural details and sharpness compared to real medical images, they successfully capture essential anatomical morphology and modality-specific characteristics. The strong downstream performance, including segmentation tasks detailed in the Appendix~\ref{app:segresults}, validates that such structurally coherent synthetic data, despite missing low-level details, is highly competent for knowledge retention in CL. This confirms that preserving core semantic and structural information is more critical than pixel-level fidelity for effective knowledge transfer in SSL. More synthetic examples are provided in the supplementary material.

\subsection{Ablation studies}
To validate the effectiveness of each component in InvCoSS, we conduct comprehensive ablation studies on six downstream tasks across imaging modalities that utilize inverted images for knowledge retention. Table~\ref{tab:ablation} reports quantitative results with different component combinations. The combination of $\mathcal{L}_{task}$ and $\mathcal{L}_{norm}$ is indispensable for successful image inversion, then we systematically analyze the effectiveness of each component.

\begin{figure}[htbp]
    \centering
    \includegraphics[width=\linewidth]{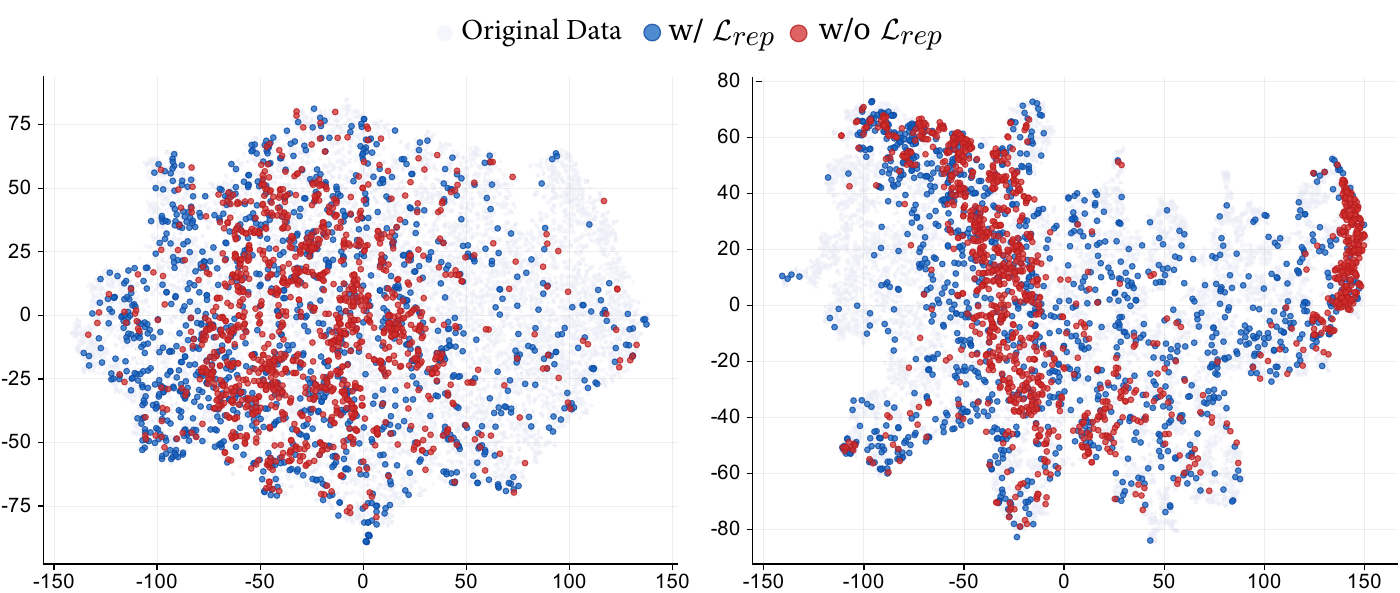}
    \caption{t-SNE visualization of real and synthetic image feature distributions. Left: X-ray. Right: Path. Gray: original data. Blue/Red: synthetic images with/without $\mathcal{L}_{rep}$. }
    \label{fig:tsne_comparison}
\end{figure}

\begin{figure}[!htbp]
    \centering
    \includegraphics[width=\linewidth]{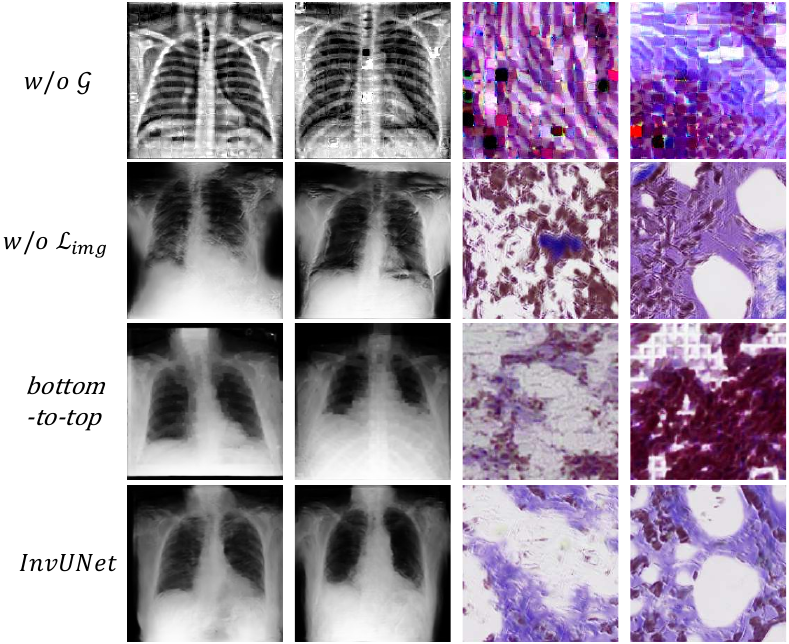}
    \caption{Visual comparison of synthetic images generated with different configurations. From top to bottom: without generator, without image prior, with bottom-to-top generator, and our InvUNet. Each row shows X-ray and pathological imaging samples.}
    \label{fig:ablation_vis}
\end{figure}

\noindent\textbf{Impact of Regularization Terms.} Adding $\mathcal{L}_{img}$ improves average performance from 86.12 to 86.75 AVG$\uparrow$ and reduces AVG$\downarrow$ from 27.12 to 24.29. As shown in Fig.~\ref{fig:ablation_vis}, removing $\mathcal{L}_{img}$ introduces high-frequency noise and local color shifts in synthetic images. This demonstrates that total variation regularization effectively suppresses artifacts and enhances spatial smoothness. Further incorporating $\mathcal{L}_{rep}$ yields the best results at 88.03 AVG$\uparrow$ and 20.11 AVG$\downarrow$, with particularly significant gains on CT and MRI tasks. On RICORD, $\mathcal{L}_{rep}$ improves accuracy from 82.14\% to 84.52\%, validating that repulsive representation learning successfully mitigates mode collapse and enhances sample diversity. To further illustrate this effect, we visualize the feature distributions using t-SNE on X-ray and pathology datasets in Fig.~\ref {fig:tsne_comparison}. $\mathcal{L}_{\mathrm{rep}}$ prevents synthetic images from clustering in high-density regions, enabling uniform coverage of the feature space and mitigating mode collapse.
\begin{table}[htbp]
   \centering
   \caption{Generator architecture comparison on X-ray tasks with model parameters and downstream performance. All models are pre-trained on 10K synthetic images.}
   \label{tab:generator_comparison}
   \renewcommand{\arraystretch}{0.85}
   \resizebox{\columnwidth}{!}{%
   \small
   \begin{tabular}{@{}lcccccc@{}}
       \toprule
       \multirow{2}{*}{\textbf{Generator}} & \multirow{2}{*}{\textbf{Params}} & \multicolumn{3}{c}{\textbf{ChestXR}} & \multicolumn{2}{c}{\textbf{QaTa}} \\
       \cmidrule(lr){3-5} \cmidrule(lr){6-7}
       & & ACC & AUC & F1 & DSC & HD \\
       \midrule
       w/o $\mathcal{G}$ & - & 89.12 & 97.21 & 88.45 & 74.42 & 36.70 \\
    bottom-to-top& 10.6M & 90.77 & 97.93 & 89.76 & 75.65 & 34.04  \\
      bottom-to-top-larger& 25.9M & 91.21 & 98.01 & 90.56 & 75.49 & 34.17  \\
       \textbf{InvUNet (Ours)} & 7.7M & \textbf{92.27} & \textbf{98.47} & \textbf{91.68} & \textbf{76.23} & \textbf{32.90}  \\
       \bottomrule
   \end{tabular}
   }
\end{table}

\noindent\textbf{Generator Analysis.} The InvUNet generator $\mathcal{G}$ is critical for computational efficiency, particularly for 3D medical imaging. As shown in the last row of Table~\ref{tab:ablation}, directly optimizing image-sized tensors without the generator becomes computationally prohibitive for 3D modalities due to the massive parameter space of volumetric data. While the baseline without $\mathcal{G}$ achieves 92.24\% ACC on 2D ChestXR tasks, it cannot scale to 3D CT and MRI tasks. To further validate our InvUNet design, we compare it against alternative generators. Table~\ref{tab:generator_comparison} shows results where models are pre-trained solely on 10,000 synthetic images and evaluated on downstream tasks. We compare against the commonly used bottom-to-top CNN in inversion at 10.6M parameters. and its larger variant at 25.9M parameters. Despite having only 7.7M parameters, InvUNet achieves the best performance at 92.27\% ACC and 98.47\% AUC on ChestXR, and 76.23\% DSC on QaTa. Fig.~\ref{fig:ablation_vis} shows that directly optimized images exhibit severe grid-like artifacts, while $\mathcal{G}_{CMI}$ loses fine-grained tissue textures. InvUNet generates the most anatomically coherent images, confirming its effectiveness for knowledge retention.

\subsection{Discussions}

\begin{figure}[htbp]
    \centering
    \includegraphics[width=\linewidth]{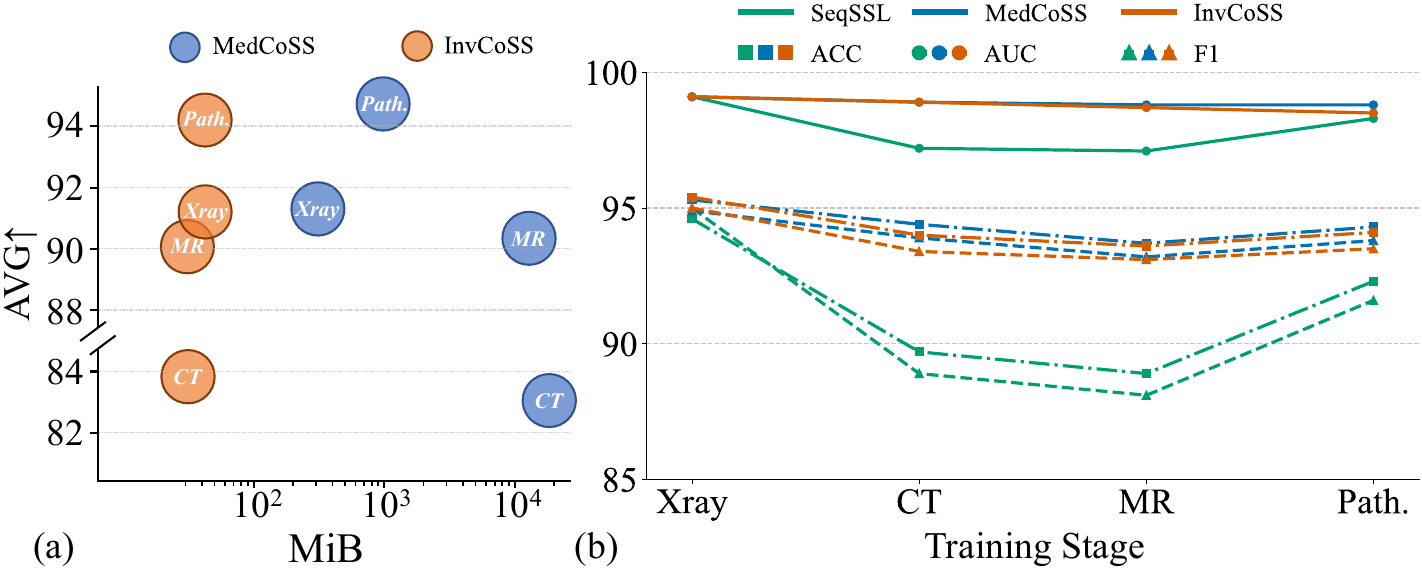}
    \caption{Storage efficiency and knowledge retention analysis. (a) Storage overhead versus average downstream performance across four modalities. (b) Performance on the ChestXR downstream task at different stages of continual learning.}
    \label{fig:storage}
\end{figure}

\noindent\textbf{Storage Efficiency.}
As illustrated in Fig.~\ref{fig:storage}(a), we compare the storage overhead with the data-replay method MedCoSS~\cite{ye2024continual}. Results for X-ray, CT, and MRI follow the default training order, while Path. uses the alternative order. We validate the robustness of our method to different modality orderings in Appendix~\ref{app:order}. All storage costs are calculated without compression. Our method achieves substantial storage reduction across all modalities, reducing storage overhead by 7$\times$ to 590$\times$ compared to MedCoSS. Despite its extremely low storage cost, our method delivers performance comparable to MedCoSS and even surpasses it on CT tasks, demonstrating that lightweight normalization statistics effectively preserve learning performance.

\begin{figure}[htbp]
    \centering
    \includegraphics[width=\linewidth]{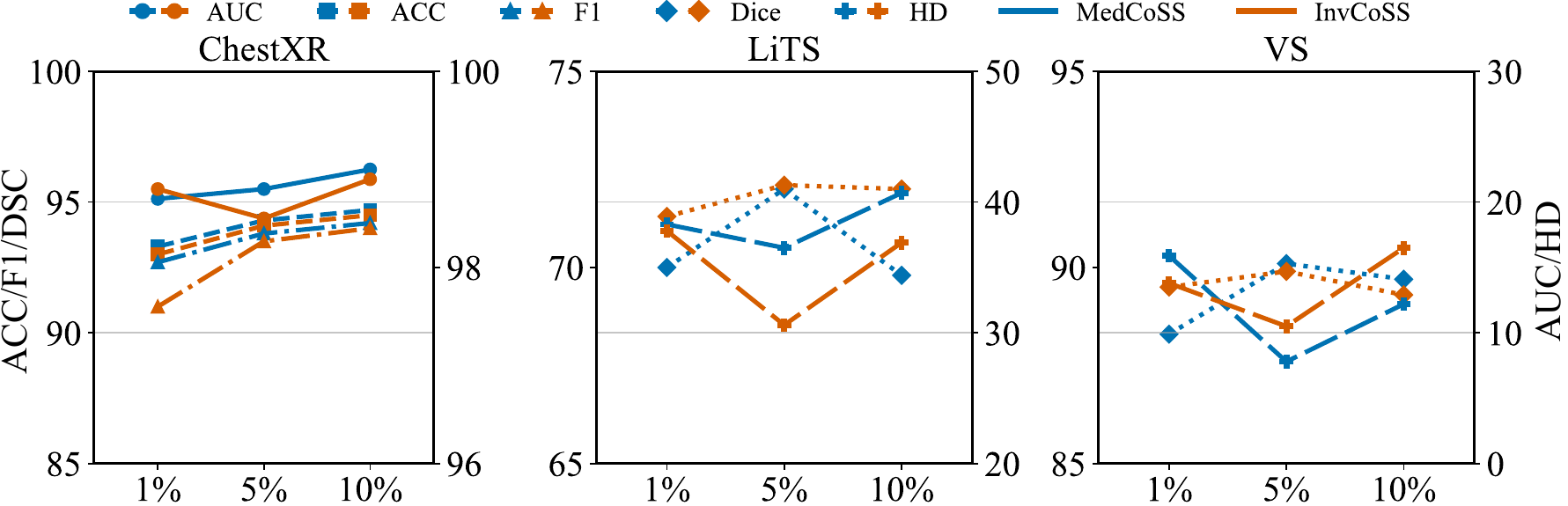}
    \caption{Performance comparison between MedCoSS and InvCoSS with 1\%, 5\%, and 10\% sample sizes.}
    \label{fig:ratio}
\end{figure}

\noindent\textbf{Knowledge Retention.}
As illustrated in Fig.~\ref{fig:storage}(b), we compare the performance of sequential pre-training (SeqSSL), MedCoSS~\cite{ye2024continual}, and our method at different training stages on the ChestXR downstream task under the default training order. The results show that SeqSSL suffers from severe catastrophic forgetting, while both MedCoSS and InvCoSS effectively maintain stable performance across all stages. This demonstrates that, under realistic data transfer constraints, our inversion-based synthetic images achieve knowledge retention comparable to that of real data replay.
\noindent\textbf{Impact of Synthetic Sample Size.}
We evaluate performance with different buffer sizes of 1\%, 5\%, and 10\% on ChestXR, LiTS, and VS tasks. As shown in Fig.~\ref{fig:ratio}, both methods demonstrate similar trends, consistent with MedCoSS's observation that larger buffers with either real or synthetic data can intensify modality conflicts and increase computational costs, hindering current modality learning. Notably, our method achieves better performance at 1\% buffer size, while matching MedCoSS at 5\% and 10\%. This suggests that with limited samples, inverted synthetic images may better capture the original data distribution than cluster-based subsets, as clustered samples may inadequately represent data diversity at small buffer sizes.

\noindent\textbf{Data Privacy Analysis.}
Despite the high fidelity of synthetic images generated via model inversion, data privacy remains well preserved. We provide a detailed visual comparison of synthetic and real images in Appendix~\ref{app:syn2real}. Compared to real images, the synthetic versions exhibit significantly less high-frequency detail across the entire image. We argue that such images are particularly suitable for SSL models, which focus on capturing global semantic information rather than local details. This characteristic not only benefits the generalization capability of downstream tasks but also alleviates data privacy concerns, especially since raw images are not stored after each pre-training stage. In summary, the high fidelity coupled with the absence of fine-grained details ensures effective representation learning in SSL, while simultaneously preventing potential data privacy leakage.

%% file: sec/5_conclusion.tex
\section{Conclusion}
In this paper, we introduce InvCoSS, a novel inversion-driven continual self-supervised learning framework that tackles catastrophic forgetting and data privacy in medical multi-modal image pre-training. By leveraging model inversion to reconstruct prior data distributions from only model checkpoints and normalization statistics, our method eliminates raw data storage while retaining learned knowledge effectively. We introduce a novel InvUNet network with multi-scale fusion to recover high-frequency details and a repulsive representation learning mechanism to enhance feature diversity. Extensive experiments on four imaging modalities and nine downstream tasks show that InvCoSS matches or surpasses data-replay performance, while reducing storage overhead by up to 590 times and completely avoiding privacy risks. This work establishes a privacy-preserving continual self-supervised learning paradigm for medical imaging, paving the way for scalable and ethical AI systems that continuously adapt to evolving clinical data while maintaining robust performance across diverse applications.

%% file: sec/appendix.tex
\clearpage
\setcounter{page}{1}
\maketitlesupplementary
\appendix
\section{Datasets}
\label{app:ds}
\subsection{Pre-training}

\textbf{MIMIC-CXR 2.0.0 Dataset}: The MIMIC-CXR 2.0.0~\cite{johnson2019mimic} dataset is a large-scale, publicly available collection of chest X-ray images. It contains a total of 377,110 JPEG chest radiographs, which are paired with 227,827 clinical reports. Following the standard preprocessing steps described in \cite{wang2022multi, zhang2022contrastive}, all lateral-view chest X-ray images were excluded, as downstream tasks focus solely on frontal-view chest radiographs. After this filtering process, the dataset was reduced to a total of 356,309 frontal-view chest X-ray images. This dataset is commonly used for tasks such as disease classification, image analysis, and other chest X-ray-related downstream applications.

\noindent{\textbf{DeepLesion dataset}}: This dataset comprises 10,594 CT scans from 4,427 participants. We standardized all scans to 1.0 $\times$ 1.0 $\times$ 1.0 mm spacing, consistent with the methodology in~\cite{ye2024continual}. To generate training samples, we divided each scan into smaller volumes using a sliding window approach, with window sizes of 24 slices and 12-slice intervals along the axial direction. These extracted segments were then normalized to dimensions of 16 $\times$ 192 $\times$ 192 voxels, yielding a total of 125,070 sub-volumes for analysis. Intensity values were normalized by mapping the Hounsfield unit range of [-1024, 1024] to [-1, 1].

\noindent{\textbf{ADNI dataset}}: We compiled our ADNI dataset by combining scans from the ADNI-1, ADNI-2, and ADNI-GO collections. Subject inclusion was determined purely by diagnostic classification (Alzheimer's disease, mild cognitive impairment, or normal cognition), with no restrictions on patient demographics or scanner specifications. From each MRI scan, we generated multiple sub-volumes through a segmentation process that extracted 16-slice windows with a stride of 8 slices along the depth direction. After resizing these segments to standardized dimensions of 16 $\times$ 192 $\times$ 192 voxels, we obtained 59,205 training samples. Intensity normalization was performed by standardizing each volume to zero mean and unit variance.

\noindent{\textbf{TCGA dataset}}: This collection encompasses histopathological images across seven tumor types from The Cancer Genome Atlas: THYM, THCA, BRCA, UCEC, UVM, OV, and MESO. We processed the whole-slide images by extracting contiguous 512 $\times$ 512 patches and resizing them to 224 $\times$ 224 resolution. From each patient's collection of $n$ patches, we randomly extracted a maximum of 100 patches for inclusion in the training set.
\subsection{Downstream}
\noindent{\textbf{PubMed20k dataset}}: We utilized a collection of 20,000 RCT abstracts containing 240,000 sentences and a vocabulary of 68,000 words. Each sentence requires assignment to one of five structural labels (background, objective, method, result, or conclusion). Following the established official data split protocol, we used only the raw text sequences for training and evaluating our sentence classification model.

\noindent{\textbf{ChestXR Dataset}}:
The ChestXR~\cite{Akhloufi2021} dataset is designed for the classification of chest X-ray images into three categories: COVID-19, pneumonia, or normal. It consists of an official split of 14,958 training images and 3,432 test images. To create a validation set, 20\% of the training images were randomly selected. 

\noindent{\textbf{QaTa-COV19-v2 (QaTa) Dataset}}:
The QaTa-COV19-v2~\cite{degerli2022osegnet} dataset is tailored for COVID-19-infected region segmentation in chest X-ray images. It provides an official split of 7,145 training images and 2,113 test images. Similar to~\cite{Akhloufi2021}, 20\% of the training data is randomly sampled to serve as the validation set. 

\noindent{\textbf{RICORD Dataset}}:
The RICORD dataset is a collection of 330 chest CT scans designated for binary classification tasks distinguishing COVID-19 positive cases from healthy controls. Each volumetric scan was standardized to 64 $\times$ 192 $\times$ 192 voxel resolution during preprocessing. 

\noindent{\textbf{RICORD Dataset}}:
The RICORD dataset is a collection of 330 chest CT scans designated for binary classification tasks distinguishing COVID-19 positive cases from healthy controls. Each volumetric scan was standardized to 64 $\times$ 192 $\times$ 192 voxel resolution during preprocessing. 

\noindent{\textbf{Vestibular Schwannoma (VS) Dataset}}:
The VS dataset provides 242 multi-sequence MRI scans (T2 and T1CE) for segmentation of vestibular schwannoma tumors. In this study, we selected the T1 contrast-enhanced sequence as input for our segmentation model. Data splits followed the protocol established in~\cite{ye2024continual}, and preprocessing was performed using the nnU-Net framework.

\noindent{\textbf{LA Dataset}}:
The LA dataset comprises 100 gadolinium-enhanced cardiac MRI scans with expert annotations for left atrium segmentation. We adopted both the data splits and preprocessing protocol from~\cite{yu2019uncertainty}.

\noindent{\textbf{NCH Dataset}}:
The NCH dataset combines the NCT-CRC-HE-100K dataset for training and the CRC-VAL-HE-7K dataset for testing. To establish a validation set, we randomly sampled 20\% of the training data from each class category. 

\noindent{\textbf{GlaS Dataset}}:
The GlaS dataset comprises 165 H\&E-stained histopathological images from colon tissue sections, annotated as malignant or benign cases. We followed the official train-test partition and created a validation set by randomly selecting 20\% of training samples from each diagnostic category. 

\section{Methodology Details of InvCoSS}

\subsection{Workflow of InvCoSS}
\label{app:invcoss}

In this section, we provide a comprehensive algorithmic description of the proposed InvCoSS framework. The core objective is to enable continual self-supervised learning without retaining raw data from previous tasks. The overall procedure, summarized in Algorithm~\ref{alg:invcoss}, consists of two distinct phases: Data-Free Model Inversion and Continual Training.

\begin{algorithm}[t]
\caption{InvCoSS Algorithm}
\label{alg:invcoss}
\small
\textbf{Input:} Frozen model $f_{T-1}$, Saved statistics $\{\mu_t, \sigma_t\}_{t=1}^{T-1}$, Current data $\mathcal{D}_T$. \\
\textbf{Output:} Updated model $f_T$.

\begin{algorithmic}[1]
\STATE \textbf{Stage 1: Data-Free Model Inversion}
\STATE Initialize synthetic buffer $\mathcal{B}_T = \emptyset$.
\FOR{each previous task $t = 1$ \TO $T-1$}
    \STATE Retrieve saved feature statistics $\{\mu_t, \sigma_t\}$.
    \FOR{each inversion iteration}
        \STATE Update $\mathcal{G}, z$ to minimize $\mathcal{L}_{\text{Inv}}$ on $f_{T-1}$ guided by $\{\mu_t, \sigma_t\}$ (Eq.~\ref{eq:generator_loss}).
    \ENDFOR
    \STATE Generate $\mathcal{D}_t^{syn}$ and update $\mathcal{B}_T \leftarrow \mathcal{B}_T \cup \mathcal{D}_t^{syn}$.
\ENDFOR

\STATE \textbf{Stage 2: Continual Training}
\STATE Initialize $f_T \leftarrow f_{T-1}$.
\FOR{each training step}
    \STATE Update $f_T$ using batches $x \sim \mathcal{D}_T$ and $x^{syn} \sim \mathcal{B}_T$ to minimize Eq.~\ref{eq:ssl_loss}.
\ENDFOR
\RETURN $f_T$
\end{algorithmic}
\end{algorithm}

\noindent\textbf{Stage 1: Data-Free Model Inversion} (Lines 1-9).
To prevent catastrophic forgetting, we construct a synthetic sample set $\mathcal{B}_T$ that approximates the data distributions of all prior tasks $t \in \{1, \dots, T-1\}$. Unlike traditional methods that require storing previous task samples, InvCoSS relies solely on lightweight feature statistics $\{\mu_t, \sigma_t\}$ stored for each task and the current pre-trained model $f_{T-1}$.
As shown in Lines 3-8, for each historical task $t$, we optimize the generator $\mathcal{G}$ and latent codes $z$ to synthesize images. Crucially, the optimization is guided by matching the stored statistics $\{\mu_t, \sigma_t\}$ while passing gradients through the frozen backbone $f_{T-1}$ (Eq.~\ref{eq:generator_loss}). This design significantly reduces storage overhead while effectively recovering domain-specific features.

\noindent\textbf{Stage 2: Continual Training} (Lines 10-14). 
Once the synthetic set $\mathcal{B}_T$ is populated, we proceed to train the current model $f_T$. We initialize $f_T$ with the weights of $f_{T-1}$ to facilitate knowledge transfer. During training, rather than relying on raw data retention, we implement a data-free knowledge distillation strategy. Specifically, each training iteration utilizes a composite mini-batch comprising real data from the current task $\mathcal{D}_T$ and synthetic samples from $\mathcal{B}_T$. The model is updated by minimizing a joint objective (Eq.~\ref{eq:ssl_loss}) that combines the task-specific loss for new knowledge acquisition and a Knowledge Distillation loss on synthetic samples for knowledge preservation.

\subsection{Batch-wise Statistics}
\label{app:mi}
Conventional model inversion techniques often exploit the running statistics of Batch Normalization (BN) layers. However, this strategy is incompatible with modern BN-free architectures like Vision Transformers (ViTs). To address this, we compute batch-wise feature statistics at the end of the final training epoch to guide the inversion process. We perform a single forward pass of the trained model $f_t$ over its entire dataset $\mathcal{D}_t$ in evaluation mode. Statistics are extracted from the feature maps of the Transformer encoder blocks. For a feature tensor of shape $(B, L, D)$, we compute the mean and variance across the batch dimension $B$ over the entire dataset, capturing the feature distribution characteristics essential for synthesizing images with coherent spatial structures. According to~\cite{welford1962note}, we update the statistics matrices ${\mu}$ and ${\sigma}^2$ to obtain batch-wise statistics per-layer in Eq.~\ref{eq:bn_loss} over mini-batches as follows:

\begin{equation}
\small
\label{eq:gls_recursive_unified}
\left( \mathbb{\mu}, \sigma^2 \right) =
\begin{cases}
   \left( \boldsymbol{\mu}_1, \boldsymbol{\sigma}^2_1 \right) & \text{if } b = 1 \\
   \begin{alignedat}{2}
       \Bigg( & \frac{N_{b-1} \boldsymbol{\mu}^{(b-1)} + n_b \boldsymbol{\mu}_b}{N_b}, \\
              & \frac{N_{b-1} (\boldsymbol{\sigma}^2)^{(b-1)} + n_b \boldsymbol{\sigma}^2_b}{N_b} \\
              & \quad + \frac{N_{b-1} n_b}{N_b^2} (\boldsymbol{\mu}^{(b-1)} - \boldsymbol{\mu}_b)^2 \Bigg)
   \end{alignedat}
   & \text{if } b > 1
\end{cases}
\end{equation}
where $N_b$ denotes the cumulative sample count up to batch $b$. The aggregated batch-wise statistics are then used as targets in a feature-matching loss, constraining the generator to produce images that replicate the learned feature distributions of the original task.

\subsection{Detail Architecture of InvUNet}
\label{app:invunet}
\begin{figure*}
    \centering
    \includegraphics[width=0.75\textwidth]{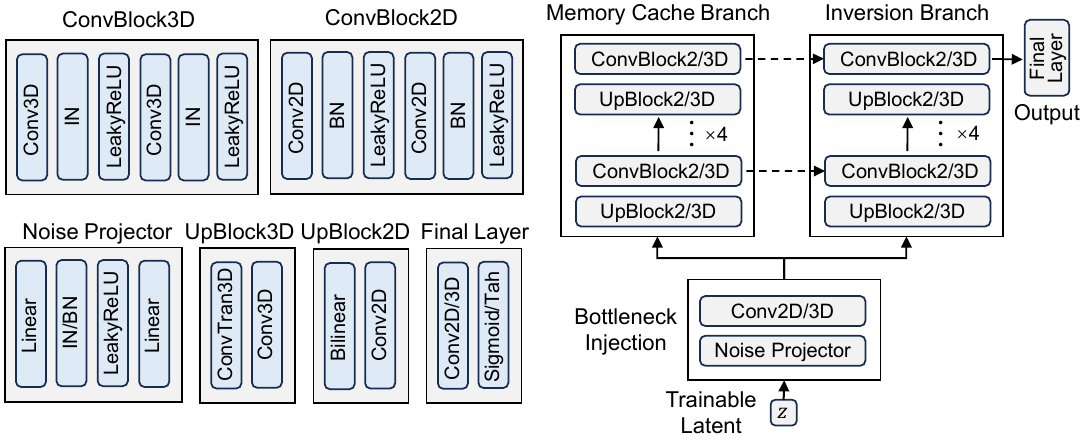}
    \caption{InvUNet architecture overview. Left panels show building blocks: ConvBlock3D and ConvBlock2D consist of two consecutive convolutions with normalization and activation; UpBlock performs 2× upsampling followed by ConvBlock; Noise Projector maps latent vector to bottleneck features; Final Layer produces output. Right side shows the dual-stream architecture with Bottleneck Injection connecting to Memory Cache Branch and Inversion Branch through skip connections.}
    \label{fig:invunet-a}
\end{figure*}

InvUNet is a dual-stream U-Net-based generator designed for high-fidelity image inversion from low-dimensional latent codes. As shown in Fig.~\ref{fig:invunet-a}, the architecture strategically injects the noisy latent vector $z$ at the network bottleneck, forcing it to encapsulate essential semantic information while delegating spatial upsampling to specialized branches. Two complementary pathways operate in parallel: the Memory Cache Branch generates multi-scale structural priors, while the Inversion Branch focuses on semantically guided, high-fidelity reconstruction through skip connections.

\noindent{\textbf{Bottleneck Injection.}} Rather than injecting $z$ at the input layer, InvUNet injects it at the network bottleneck through a Noise Projector module. The latent vector is first linearly projected to match the bottleneck feature map size, then passed through normalization and activation. This design forces the latent representation to encapsulate essential semantic information and prevents direct high-dimensional upsampling of raw noise.

\noindent{\textbf{Memory Cache Branch.}} The Memory Cache Branch operates as a lightweight encoder that generates multi-scale structural priors. Starting from the bottleneck, it progressively upsamples features through successive UpBlock modules. Each UpBlock performs $2\times$ upsampling followed by ConvBlock operations. For 2D images, the branch uses bilinear interpolation and 2D convolutions with Batch Normalization. For 3D volumes, it employs trilinear interpolation and 3D convolutions with Instance Normalization to maintain batch independence. The structural priors generated at each level are stored and later used as skip connections to guide the Inversion Branch.

\noindent{\textbf{Inversion Branch.}} The Inversion Branch is the primary reconstruction pathway that focuses on high-fidelity semantic-guided synthesis. It also progressively upsamples through UpBlock and ConvBlock modules, but additionally receives skip connections from the Memory Cache Branch at each corresponding resolution level. These skip connections enable the Inversion Branch to access structural guidance while focusing its capacity on refining semantic details and high-frequency information. The interplay between upsampling and skip-guided feature fusion ensures effective multi-scale reconstruction.

\noindent{\textbf{Final Layer.}} After reconstruction at full resolution, the features are processed through a final output layer. For 2D image synthesis, this consists of a Conv2d layer with Sigmoid activation to produce images in the range $[0, 1]$. For 3D volumetric synthesis, the final layer consists of a Conv3d layer with Tanh activation to produce volumes in the range $[-1, 1]$, which is more suitable for medical imaging applications.

\noindent{\textbf{Configuration Examples.} InvUNet can be applied to both 2D image and 3D volume medical image synthesis. The following Table~\ref{tab:invunet_config} demonstrates the architecture for both modalities.

\begin{table}[h]
\centering
\small
\caption{Architectural configurations of InvUNet for both 2D and 3D synthesis tasks, producing outputs at $224 \times 224$ resolution for 2D images and $16 \times 192 \times 192$ resolution for 3D volumes.}
\label{tab:invunet_config}
\resizebox{\linewidth}{!}{%
\begin{tabular}{lcccc}
\toprule
Layer & \multicolumn{2}{c}{\makecell{2D Config\\ 224×224}} & \multicolumn{2}{c}{\makecell{3D Config\\ 16×192×192}} \\
\cmidrule(lr){2-3} \cmidrule(lr){4-5}
 & \makecell{Spatial\\Resolution} & Channels & \makecell{Spatial\\Resolution} & Channels \\
\midrule
Bottleneck & 14×14 & 256 & 1×12×12 & 256 \\
\midrule
\multicolumn{5}{c}{\small \textit{Memory Cache Branch}} \\
\midrule
MC-1 & 28×28 & 128 & 2×24×24 & 128 \\
MC-2 & 56×56 & 64 & 4×48×48 & 64 \\
MC-3 & 112×112 & 32 & 8×96×96 & 32 \\
MC-4 & 224×224 & 16 & 16×192×192 & 16 \\
\midrule
\multicolumn{5}{c}{\small \textit{Inversion Branch}} \\
\midrule
Inv-1 & 28×28 & 128 & 2×24×24 & 128 \\
Inv-2 & 56×56 & 64 & 4×48×48 & 64 \\
Inv-3 & 112×112 & 32 & 8×96×96 & 32 \\
Inv-4 & 224×224 & 16 & 16×192×192 & 16 \\
\midrule
Output & 224×224 & 3 & 16×192×192 & 1 \\
\bottomrule
\end{tabular}
}
\end{table}

\section{Additional Implementation Details}
\label{app:id}
All experiments are implemented in PyTorch and conducted on a cluster with 8 NVIDIA RTX 4090 GPUs.

\noindent\textbf{Multi-modal Backbone and Pre-training.}
To accommodate diverse medical modalities, we employ a dimension-free architecture based on a standard ViT/B encoder. We use dimension-specific tokenizers to process inputs: Byte Pair Encoding for 1D clinical reports, and patch embeddings for 2D (X-rays, Pathology) and 3D (CT, MRI) visual data. For pre-training objectives, we follow modality-specific masking strategies. For text, we randomly mask 15\% of words and optimize with Cross-Entropy loss following BERT. For 2D and 3D visual data, we adopt a high masking ratio of 75\% and employ Mean Squared Error (MSE) loss on masked regions to reconstruct pixel values.

\noindent\textbf{Modality-Specific Inversion Strategy.}
We apply our inversion-based replay strategy to all imaging modalities (X-ray, CT, MRI, and Pathological imaging) encountered during CSSL. For the Report modality, we adopt the clustering-based replay method following MedCoSS~\cite{ye2024continual}, as textual data is highly compressed and poses minimal privacy or storage risks compared to medical imaging data.

\noindent\textbf{InvUNet Configurations.}
For image inversion, we employ the proposed InvUNet architecture with a channel configuration of $[16, 32, 64, 128, 256]$ for the decoder path. The specific architectural variants for 2D patches ($224\times224$) and 3D volumes ($16\times192\times192$) are detailed in Table~\ref{tab:invunet_config}.

\noindent\textbf{Hyperparameters for Synthesis.}
The optimization of the generator involves precise hyperparameter tuning to balance diversity and fidelity. \textbf{Optimization.} We use the Adam optimizer with a learning rate of $2\times10^{-4}$ for the generator parameters and a higher learning rate of $0.05$ for the trainable latent vectors $z$ (dimension 512). The loss balancing weights are set to $\alpha_{\text{norm}}=1$, $\alpha_{\text{img}}=0.1$, and $\alpha_{\text{rep}}=0.1$. \textbf{2D Synthesis Strategy.} For 2D modalities, we synthesize batches of 320 samples, optimizing each for 1,000 steps. To promote distribution diversity, we reinitialize the generator $\mathcal{G}$ and 128-dim latents $z$ after each batch generation. \textbf{3D Synthesis Strategy.} Due to higher computational costs, we generate batches of 48 samples for 3D data with 512-dim latents $z$. Optimization runs for 500 steps per batch with an extended initial generation phase of 2,000 steps. We reinitialize the generator every 10 batches to introduce variability. \textbf{Repulsive Learning.} For the repulsive term, we maintain a feature pool size matching the number of synthetic samples.
\section{Visualization}
\subsection{Qualitative Segmentation Results}
\label{app:segresults}
\begin{figure}[t]
    \centering
    \includegraphics[width=\linewidth]{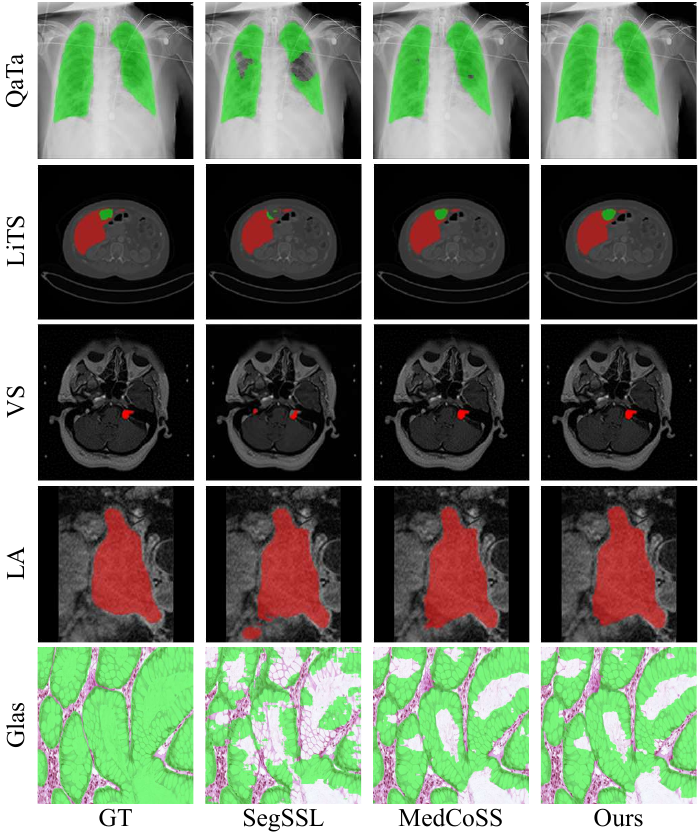}
    \caption{Visualization of segmentation results obtained by SeqSSL, MedCoSS, and Ours. Results for QaTa, LiTS, LA, and VS follow the
    default training order, while Glas uses the alternative order. }
    \label{fig:app_seg}
\end{figure}
To intuitively demonstrate the effectiveness of InvCoSS in alleviating catastrophic forgetting, we visualize the segmentation predictions on five downstream tasks in Fig.~\ref{fig:app_seg}. As observed, the sequential training baseline (SeqSSL) suffers from significant performance degradation, failing to delineate anatomical boundaries accurately. In contrast, InvCoSS produces precise segmentation masks that capture fine structural details, achieving qualitative performance highly comparable to the data-replay baseline MedCoSS. This consistency holds across different modality orders, verifying that our generated synthetic data successfully retains the necessary semantic knowledge for pixel-level dense prediction tasks.

\subsection{Visual comparison of synthetic and real image}
\label{app:syn2real}
Fig.~\ref{fig:app_privacy} provides a close-up comparison between real and synthetic X-ray images to illustrate the privacy-preserving properties of our method. As highlighted by the blue boxes, real medical images contain distinct high-frequency details, including clear rib textures, bone structures, subtle lesions, and medical instruments. In contrast, our synthetic images, while preserving the global anatomical layout (e.g., lung fields and spine) essential for representation learning, lack these fine-grained, patient-specific specificities. This visual evidence supports our analysis that InvCoSS inherently mitigates privacy risks by discarding sensitive high-frequency details while retaining the semantic information required for effective SSL pre-training.
\begin{figure}[htbp]
    \centering
    \includegraphics[width=\linewidth]{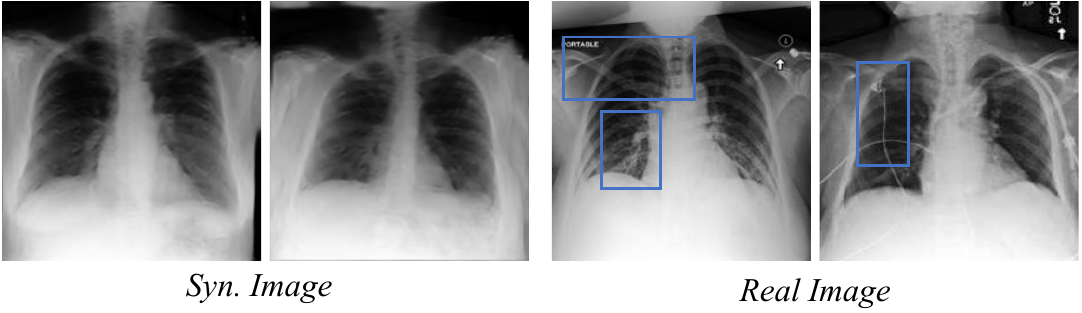}
    \caption{Visualization of real and synthetic images. }
    \label{fig:app_privacy}
\end{figure}
\section{Pre-training Order of Modality}
\label{app:order}
\begin{table}[htbp]
\centering
\caption{Performance comparison of different pre-training data orderings. The default order is: Report, X-ray, CT, MRI, Path., and the alt order is: Path., MRI, Report, CT, X-ray.}
\label{tab:results}
\small
\setlength{\tabcolsep}{1.5pt}
\renewcommand{\arraystretch}{0.75}
\resizebox{\columnwidth}{!}{
\begin{tabular}{cc|ccc|ccc}
\toprule
\multicolumn{2}{c|}{\multirow{2}{*}{Metrics}} & \multicolumn{3}{c|}{\textit{Default Order}} & \multicolumn{3}{c}{\textit{Alt Order}} \\
\cmidrule(lr){3-5} \cmidrule(lr){6-8}
& & SeqSSL & MedCoSS & InvCoSS  & SeqSSL & MedCoSS & InvCoSS \\
\midrule
\multirow{3}{*}{Pub.} & ACC & 82.12 & 83.59 & 83.63& 82.60 & 83.63 & 83.78 \\
& AUC & 94.91& 95.38 & 95.24 & 94.83& 95.27 & 95.32 \\
& F1 & 76.37& 77.87 & 77.81 & 76.52& 77.79 & 77.83  \\
\midrule
\multirow{3}{*}{Che.} & ACC & 92.65& 94.31 & 94.09 & 95.57& 95.70 & 95.50\\
& AUC &97.75& 98.83 & 98.52 & 99.10& 99.25 & 99.17 \\
& F1 & 92.01& 93.77 & 93.53 & 95.08& 95.24 & 95.08 \\
\midrule
\multirow{2}{*}{LiTS} & DSC & 66.57 & 72.01 & 72.14 & 70.60 & 71.43 & 71.22\\
& HD & 49.42 & 36.50 & 30.36 & 42.27 & 36.58 & 38.75 \\
\midrule
\multirow{2}{*}{VS} & DSC & 86.29 & 90.12 & 89.93 & 83.76 & 89.29 & 88.60 \\
& HD & 34.90 & 7.80 & 10.25 & 62.23 & 11.93 & 15.22 \\
\midrule
\multirow{3}{*}{NCH} & ACC & 95.83 & 95.76 & 95.75 & 92.42 & 93.75 & 93.42 \\
& AUC & 99.57 & 99.51 & 99.61 & 99.16 & 99.30 & 99.23  \\
& F1 & 94.37 & 94.01 & 93.90 & 89.87 & 91.19 & 90.61 \\
\bottomrule
\end{tabular}
}
\end{table}
To investigate the robustness of InvCoSS to different modality ordering, we evaluate performance under two pre-training sequences: the default order of Report, X-ray, CT, MRI, Path., and an alternative order of Path., MRI, Report, CT, X-ray. Table~\ref{tab:results} presents results across five representative downstream tasks on SeqSSL, MedCoSS and InvCoSS. InvCoSS demonstrates consistent and robust performance across both orderings, maintaining competitive results with small performance variation.